\documentclass[letterpaper, 10 pt, conference]{ieeeconf} 

\IEEEoverridecommandlockouts 

\usepackage[pdftex]{graphicx} 
\usepackage{colortbl}
\usepackage{multirow} 
\newcommand{\thline}{\noalign{\hrule height 0.1pt}} 
\newcommand{\bhline}{\noalign{\hrule height 1.0pt}} 
\usepackage{CJKutf8}

\title{\LARGE \bf
Spoken Dialogue Strategy Focusing on Asymmetric Communication \\ with Android Robots
}

\author{Daisuke Kawakubo$^{1}$, Hitoshi Ishii$^{1}$, Riku Okazawa$^{1}$, Shunta Nishizawa$^{1}$, Haruki Hatakeyama$^{1}$ \\ Hiroaki Sugiyama$^{2}$, Masaki Shuzo$^{1}$ and Eisaku Maeda$^{1}$
\thanks{*This work was supported by Grant-in-Aid for Scientific Research on Innovative Areas, Grant Numbers JP19H05693.}
\thanks{$^{1}$D. Kawakubo, H. Ishii, R. Okazawa, S. Nishizawa, H. Hatakeyama, M. Shuzo and E. Maeda are with Tokyo Denki University, 5 Senju Asahi-cho, Adachi-ku, Tokyo 120-8551, Japan 
 {\tt\small \{22amj10@ms, 19aj008@ms, 19aj023@ms, 19aj110@ms, 19aj115@ms, shuzo@mail, maeda.e@mail\}.dendai.ac.jp}}
\thanks{$^{2}$H. Sugiyama is with NTT Communication Science Laboratories, 2-4 Hikaridai Seika-cho, Soraku-gun, Kyoto 619-0237, Japan 
 {\tt\small h.sugi@ieee.org}}
}

\begin{document}
\maketitle
\thispagestyle{empty}
\pagestyle{empty}

\begin{abstract}
Humans are easily conscious of small differences in an android robot's (AR's) behaviors and utterances, resulting in treating the AR as not-human, while ARs treat us as humans.
Thus, there exists asymmetric communication between ARs and humans.
In our system at Dialogue Robot Competition 2022, this asymmetry was a considerable research target in our dialogue strategy.
For example, tricky phrases such as questions related to personal matters and forceful requests for agreement were experimentally used in AR's utterances.
We assumed that these AR phrases would have a reasonable chance of success, although humans would likely hesitate to use the phrases.
Additionally, during a five-minute dialogue, our AR's character, such as its voice tones and sentence expressions, changed from mechanical to human-like type in order to pretend to tailor to customers.
The characteristics of the AR developed by our team, DSML-TDU, are introduced in this paper. 
\end{abstract}

\section{INTRODUCTION}
\begin{figure}[thpb]
  \centering
  \includegraphics[scale=1.0]{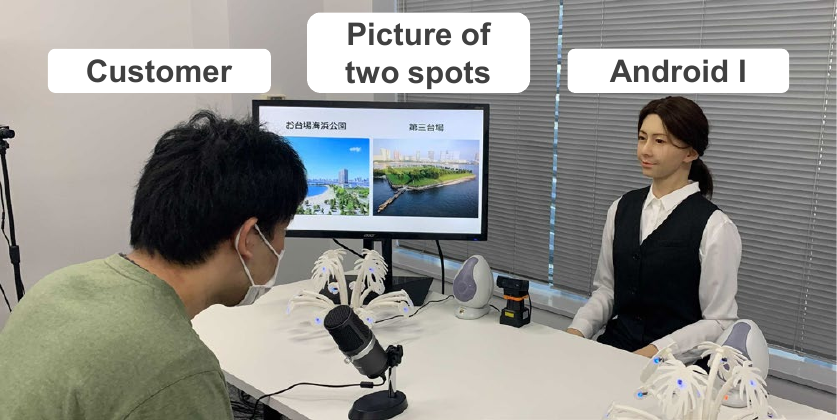}
  \caption{Android I works as counter salesperson and has dialogue with customer. Pictures of two spots are shown on the display.}
  \label{figure1}
\end{figure}

In addition to a human-like appearance, smooth movements, and rich facial expressions, the intelligence of android robots (ARs) is being improved so as to become closer to the human level in spoken dialogues.
AR systems developed by Ishiguro {\it et al}, such as the Geminoid \cite{Nishio 2007} and ERICA \cite{Glas 2016}, were made available to us participants in the Dialogue Robot Competition (DRC).
The scope of the competition was to evaluate an AR's practicality in real situations.
For the task of sightseeing-spot recommendation at a travel agency, participants tried to develop an original multimodal dialogue system using Android I, an ERICA-based platform (Fig. \ref{figure1}).
For the 2022 competition \cite{Minato 2022}, our team, DSML-TDU, which had joined in the first competition in 2020 \cite{Higashinaka 2022}, updated the system, focusing on asymmetric communication \cite{Kawamoto 2022} as explained below.

Humans are easily conscious of small differences in an AR's behaviors and utterances.
Therefore, we tend to treat ARs nowadays as not-human, although our developed ARs treat us as humans.
This asymmetricity can be seen when we communicate with an AR.

We should try to develop dialogue strategies in consideration of this asymmetric communication between robots and humans, as Bono \cite{Bono 2015} also said ``we can only develop conversations between robots and humans on the basis of the `differences' that humans unconsciously recognize as species.'' 
We assumed that ARs would have acceptable communication styles which real humans would probably avoid using.
In the case of communicating with someone for the first time, for example, humans will talk starting with casual topics and avoid personal matters.
Here, we have a question whether we can allow the AR's impoliteness or lacking consideration. 
As a discussion of this asymmetric communication, our team's system and competition results at DRC 2022 are described in this paper.

\begin{figure*}[thpb]
  \centering
  \includegraphics[scale=1.0]{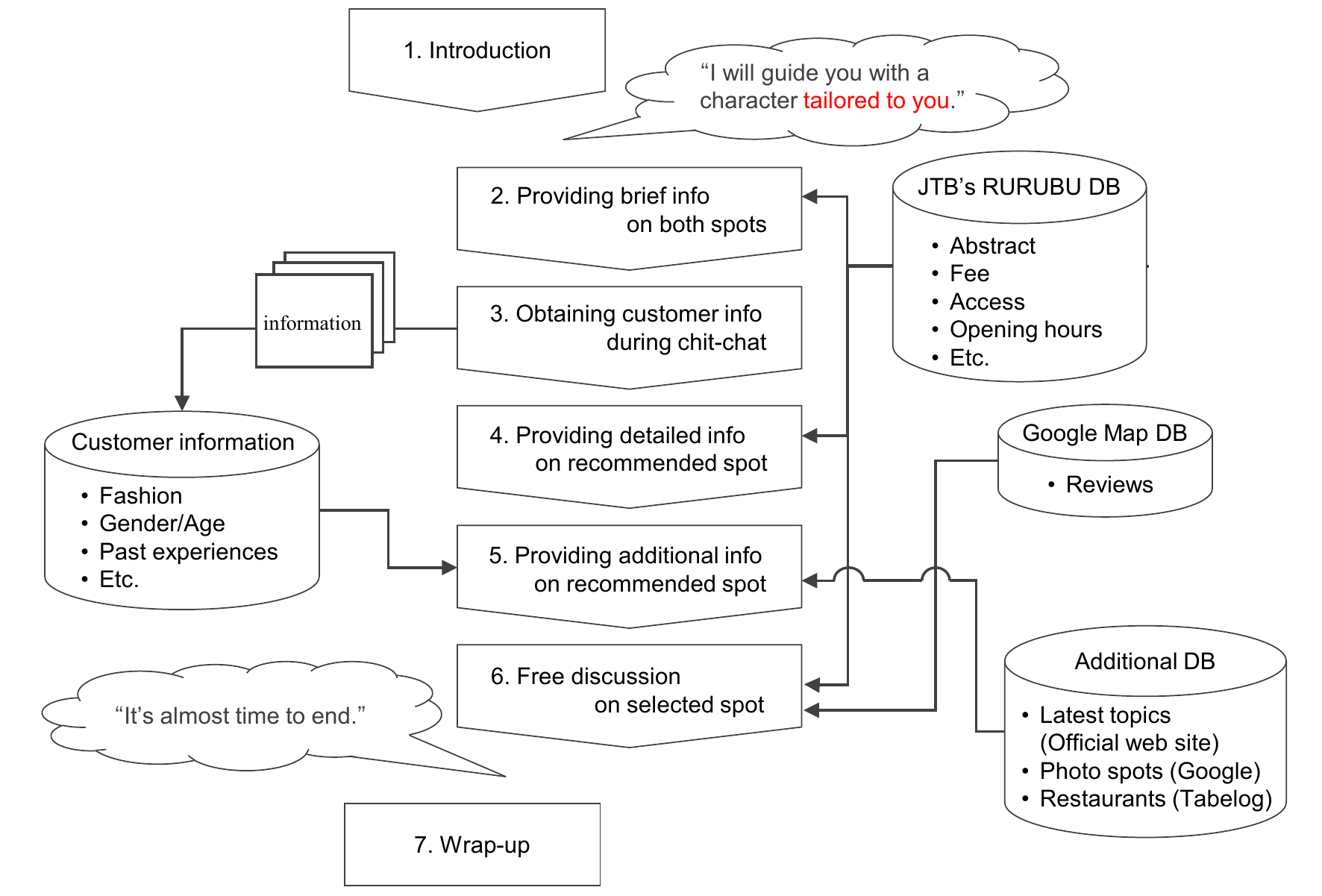}
  \caption{Dialogue flow of proposed system.}
  \label{figure2}
\end{figure*}

\begin{figure*}[thpb]
 \centering
 \includegraphics[scale=1.0]{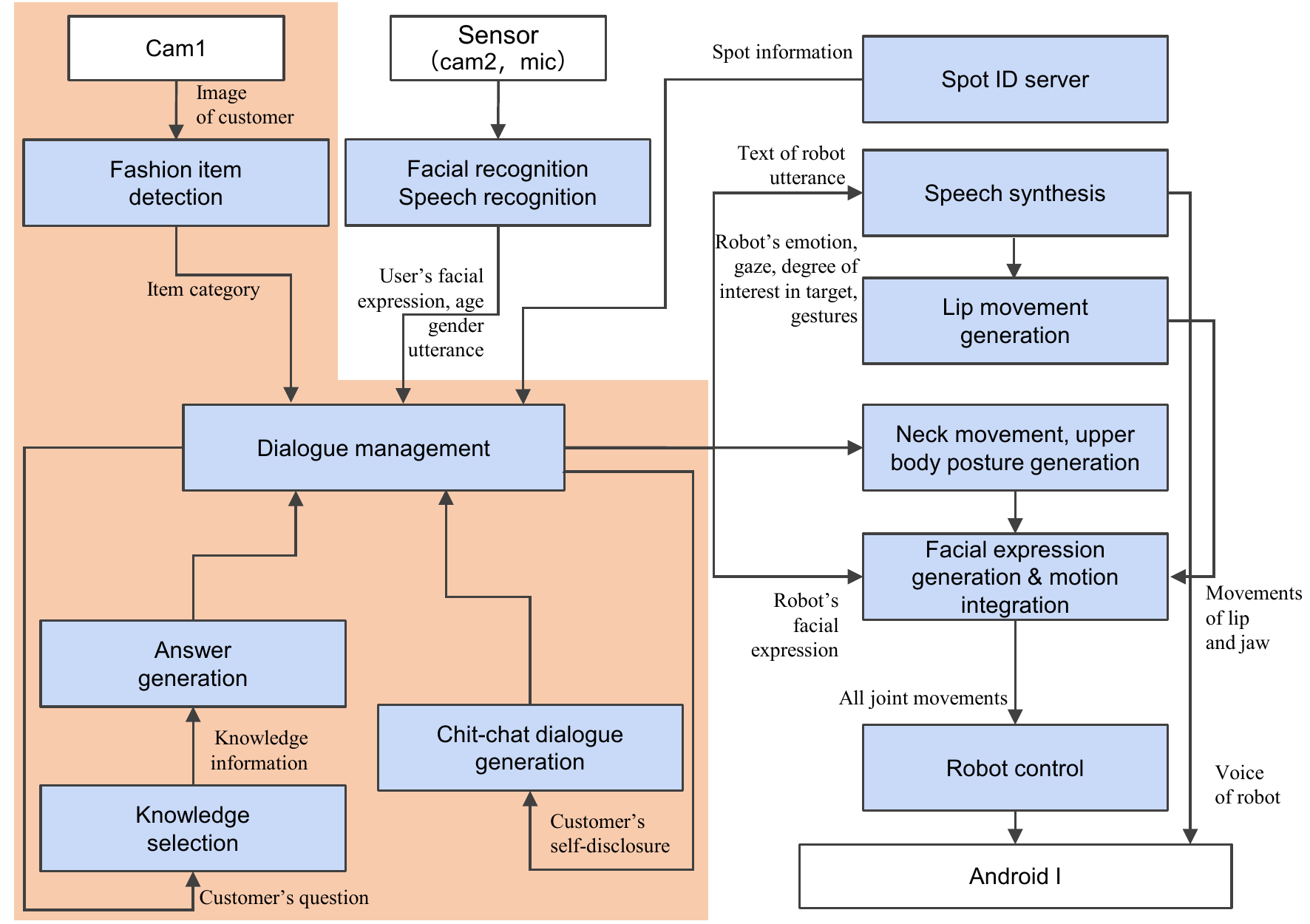}
 \caption{Configuration of system by Team DSML-TDU.
 Our developed modules are in the frame filled in salmon pink.
 They were implemented into android robot system prepared by competition organizers.
 The AR used multimodal inputs from customer to communicate with him/her.}
 \label{figure3}
\end{figure*}

\section{OUR TASKS IN DRC 2022}
The preliminary round of DRC 2022 was held at the mock travel agency booth in Miraikan.
The detail of the competition are shown in the overview paper \cite{Minato 2022}.
Android I acts as a salesperson at a travel agency and communicate with a customer.
A customer as an experimental participant who has two alternative spots selected previously by him/herself decides one through a 5-minute dialogue with Android I.
In this situation, our challenging tasks are
to give a customer enough information for both spots,
to make him/her feel enjoyable through the dialogue,
and to lead his/her decision to the recommended spot randomly designated by organizers.

\section{DIALOGUE STRATEGY FOCUSING ON ASYMMETRIC COMMUNICATION}
The dialogue of the proposed system consisted of 7 phases (Fig. \ref{figure2}).
Our system consists of a dialogue transition control system, a fashion item detection module, a chit-chat dialogue generation module, a knowledge selection module, and an answer generation module (Fig. \ref{figure3}).



ARs have a human-like appearance, so their behaviors that deviate greatly from humans can cause customers to feel uncomfortable.
This is important in focusing on asymmetric communication.
We used tricky phrases in the proposed system.
If these phrases were used by a human, the customer would feel uncomfortable, but if they are used by an AR, the customer could see them as acceptable and have a good impression.

\subsection{Shortening Psychological Distance from Customer}
\label{fashion}
At the beginning of dialogue, the customer expects the AR to make utterances and has an attitude to listen.
Therefore, from the beginning, the AR uses the customer's fashion item to say, ``Your glasses are very nice.''
If a human made this utterance in a first meeting with a customer, the customer would feel uncomfortable as this would be too personal, but if an AR makes this utterance, it can shorten the psychological distance with him/her from the beginning of dialogue.

\subsection{Natural Leading Without Feeling Intentional}
\label{pressure}
After providing information on a recommended spot to the customer, the AR asks, ``Does today's dialogue make you want to visit the recommended spot?''
The purpose of this question is to give the customer the impression that the customer had chosen the recommended spot by their own will.
If a human asks this question, the customer would feel mentally pressured.
However, since the AR asks this question, the customer would be more likely to believe that this question was not intentional, and the robot recommendation effect will increase further.

If the customer responds affirmatively to this question, the AR engages in free discussion related to the topic of the recommended spot.
In case of the negative response, or if the customer wants information on the other spot (non-recommended spot), the free discussion is then related to the topic of the non-recommended spot. 
Although providing information on the non-recommended spot would cause the robot recommendation effect to be lower, this was done to follow a competition regulation stating that the customer was to check information on both spots.

\section{PROPOSED SYSTEM}
In previous section, we discussed asymmetric communication such as tricky phrases that would be acceptable for humans.
Then, in this session, we introduce AR's behaviors and utterances aiming at human-likeness.



\subsection{Character Transforming for Pretending to Tailor to Customer}
\label{character}
\begin{figure}[bthp]
 \centering
 \includegraphics[scale=1.0]{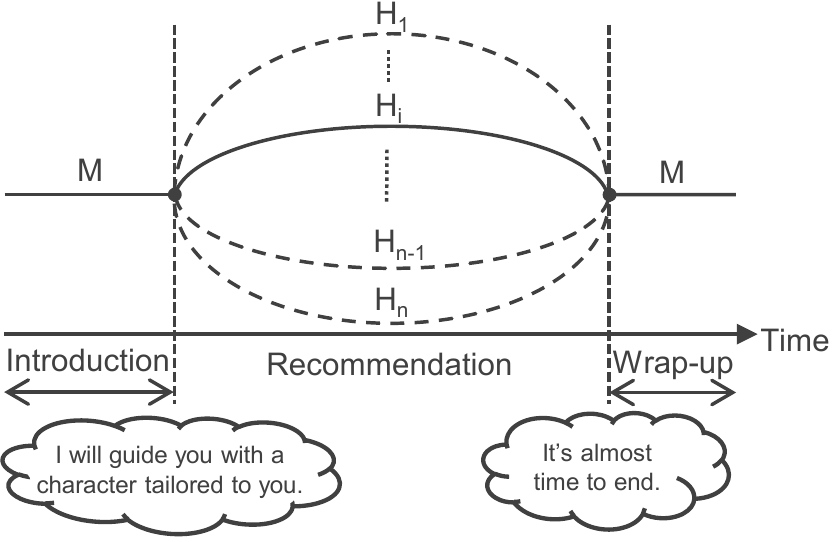}
 \caption{Transition in robot's internal states during a 5-minute dialogue. Robot takes M (machine-like) mode at beginning and end of dialogue, and H (human-like) mode between them.
 Machine-like character is transformed to human-like character so that customer can feel that AR is serving him/her.}
 \label{figure4}
\end{figure}

A dialogue system should change its response in accordance with the customer's internal state \cite{Kodama 2020}.
By clearly changing the AR's attributes, personality, and interests to be tailored to the customer, as shown in the Fig. \ref{figure4}, the customer can gain a feeling of familiarity toward the AR.
During the introduction in the dialogue, the AR says, ``I will guide you with a character tailored to you,'' and then transforms from a machine-like AR with a low voice and stiff tone of voice to a human-like AR with a high voice and soft tone of voice.


Although we could prepare several types of characters (shown in Fig. \ref{figure4}), a customer could not see others' dialogues under the competition regulation.
Therefore, the parameters of human-like and machine-like was consistently set to above mentioned condition through our preliminary round in DRC 2022 so as to analyze the experimental results easily. 
In spite of the consistency through all dialogues, we think all customers  might regard the character change as robot's serving for themselves.

\begin{figure}[bthp]
  \centering
  \includegraphics[scale=1.0]{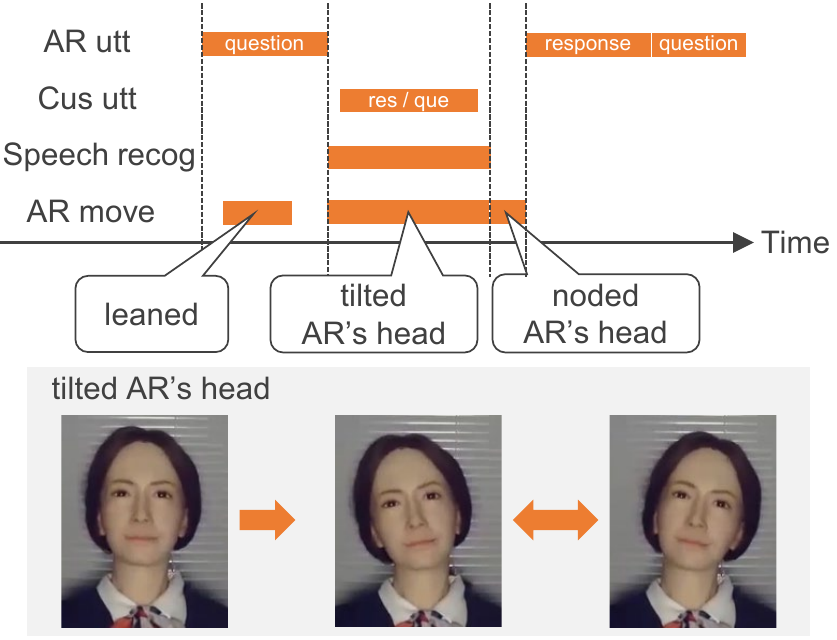}
  \caption{Transition of AR's movements for smooth turn-taking.}
  \label{figure5}
\end{figure}

\subsection{Reducing Mental Stress in Customer}
\label{movement}
Unsmooth turn-taking may cause the customer to feel uncomfortable. 
Therefore, by clearly indicating the listening state of the AR, the customer's mental stress can be reduced.
Our developed transition chart was shown in Fig. \ref{figure5}.
When the AR started to have a question, it leaned forward to indicate to the customer that turn-taking will be occurring.
During the customer answering, the AR tilted its head at regular intervals to indicate to the customer that it was listening.
After finishing the customer's utterance, the AR nodded its head deeply twice to indicate to him/her that it was recognizing the utterance while generating next AR's utterance.


\subsection{Building More Natural Dialogue}
\label{hobbyist}
If the AR only asked closed questions to customers, it could give the impression of interrogating them.
In the dialogue, the AR asked open questions during chit-chat, which allowed customers to speak freely and get a good impression of the AR (Table \ref{table1}).

The important thing in a dialogue is to share and understand each other's thoughts, so an AR should not speak and ask questions unilaterally.
The dialogue of the proposed system incorporated free discussion in which the customer's questions are answered, so we aimed to achieve this kind of dialogue (Table \ref{table2}).

The AR needs to respond naturally to the customer's free utterances.
In some cases, the AR responds to the customer's utterances using sentences generated by the model for chit-chat dialogue system \cite{Sugiyama 2021}.

\begin{CJK}{UTF8}{ipxm}
\begin{table}[thpb]
\caption{Example of the dialogue during open questions from the AR.}
\label{table1}
\begin{center}
\begin{tabular}{l|l}
\hline
Sys & \begin{tabular}[c]{@{}l@{}}
Now, I would like to ask you a few questions so that I can\\ recommend a spot!
\textbf{First, What past trips do you remember?}\\ (では，これからお薦めの観光地を提案するために\\ あなたについていくつか聞きます．
まず，あなたの\\ 思い出に残ってる旅行先はどこですか？) \end{tabular} \\ \hline
Cus  & Germany. (ドイツ) \\ \hline
Sys & \begin{tabular}[c]{@{}l@{}}
Germany sounds great!
\textbf{So, what do you remember about}\\ \textbf{Germany?} (ドイツいいですね！
じゃあ，ドイツへの\\ 旅行ではどんなことが思い出に残ってますか？\end{tabular} \\ \hline
Cus  & \begin{tabular}[c]{@{}l@{}}\textbf{}
I remember seeing castles, eating meals, and riding the train.\\ 
(ドイツにしかないお城を見たりご飯食べたりしたこと\\ とかあとは電車に乗ったこととか覚えてます)\end{tabular} \\ \hline
Sys & \begin{tabular}[c]{@{}l@{}}
I wonder what German castles look like.
\textbf{Who would you}\\ \textbf{like to go on the trip with?}(ドイツのお城どんなふうに\\ なっているのか気になります 今日考えてる旅行へは誰と\\ 行きたいですか？) \end{tabular} \\ \hline
Cus  & 
I want to go with friends. (友達と行きたいです) \\ \hline
Sys & \begin{tabular}[c]{@{}l@{}}
So you are traveling with friends. (友達との旅行なんですね．)\end{tabular} \\ \hline
\end{tabular}
\end{center}
\end{table}
\end{CJK}

\begin{CJK}{UTF8}{ipxm}
\begin{table}[thpb]
\caption{Example of the dialogue responding to customer's questions.}
\label{table2}
\begin{center}
\begin{tabular}{l|l}
\hline
Sys & \begin{tabular}[c]{@{}l@{}}
So, do you have any questions about Tokyo Trick Art Museum,\\ such as fees or the parking area?
(では，東京トリックアート\\ 迷宮館について，料金や駐車場など質問はありますか？) \end{tabular} \\ \hline
Cus  & 
\textbf{How much are the fees?}
(料金はいくらですか) \\ \hline
Sys & \begin{tabular}[c]{@{}l@{}}
High school students and older are 1,000 yen, 4 years old to\\ junior high school students are 700 yen, and children under\\ 3 years old are free.
Do you have any questions?
(高校生以上\\ 1000円、4歳から中学生が700円、3歳以下は無料です。\\ 他に質問はありますか？) \end{tabular}\\ \hline
Cus  & 
\textbf{Where is the nearest station?} (最寄り駅はどこですか) \\ \hline
Sys & \begin{tabular}[c]{@{}l@{}}
It is a 2-minute walk from Odaiba Marine Park Station on the\\ Yurikamome Line or a 3-minute drive from Daiba Gateway on\\ the Metropolitan Expressway.
(ゆりかもめ「お台場海浜公園駅」\\ 徒歩2分，または首都高速「台場出入口から3分です。) \end{tabular}\\ \hline
\end{tabular}
\end{center}
\end{table}
\end{CJK}

\begin{CJK}{UTF8}{ipxm}
\begin{table}[thpb]
        \caption{
        Impression evaluation of the dialogue and the robot recommended effect by the questionnaire survey from 29 customers.
        Baseline system is general recommendation dialogue system created by organizers.
        Sat/c, Inf, Nat, App, Lik, Sat/d, Tru, Use, Reu, and Recom denote satisfaction with choice, informativeness, naturalness, appropriateness, likeability, satisfaction with dialogue, trustworthiness, usefulness, intention to reuse, and robot recommendation effect, respectively.
        }
    \label{table3}
    \begin{center}
    \begin{tabular}{ccc}
    \bhline
    \multirow{2}{*}{Questionnaire items} & \begin{tabular}[c]{@{}c@{}}Proposed\\ system\end{tabular} & \begin{tabular}[c]{@{}c@{}}Baseline\\ system\end{tabular} \\
    & mean±SD & mean \\
    \thline
    Sat/c & 4.7±1.6 & 4.2 \\
    Inf & 5.0±1.7 & 4.0 \\
    Nat & 3.9±1.6 & 3.8 \\
    App & 4.5±1.7 & 4.4 \\
    Lik & 4.7±1.7 & 4.6 \\
    Sat/d & 5.0±1.6 & 4.1 \\
    Tru & 4.5±1.9 & 4.3 \\
    Use & 4.9±1.4 & 4.7 \\
    Reu & 4.6±1.8 & 4.1 \\
    Recom & 11.5±26.2 & 5.4 \\
    \bhline
    \end{tabular}
    \end{center}
    \end{table}

\end{CJK}

\addtolength{\textheight}{-10cm}

\section{ANALYSIS OF RESULTS}
Questionnaire results from the competition are shown in Table \ref{table3}.
The definition of detail items in this questionnaire are shown in the overview paper \cite{Minato 2022}.
In order to evaluate the asymmetric communication, we focused on three items of ``satisfaction with the dialogue (Sat/d)'', ``trustworthiness (Tru)'' and ``the robot recommendation effect (Recom)''.
The former two scores (Sat/d and Tru) of our system were higher than ones of the baseline system, showing  $5.0 \pm 1.6 > 4.1$ and $4.5 \pm 1.9 > 4.3$, respectively.
The last (Recom) which means the degree of success to lead the customer to the recommendation spot was also higher than baseline, showing $11.5 \pm 26.2 > 5.4$.
These results showed our dialogue strategy may be effective.
The other scores which seemed to have no significant difference from baseline will be discussed in the future work with other teams data.




We watched videos of dialogues with customers that had a low evaluation.
From the videos, we confirmed cases of dialogue breakdown due to errors in the speech recognition software and the fashion item detection module, as well as cases in which the speech synthesizer misread the generated sentences.

A total of 29 dialogue evaluations of our system were conducted in one day.
Looking at the questionnaire results over time, there were 9 later cases for which the results were especially low.
The experiment was conducted on a holiday, so the environment in the latter half of the day was noisy.
This may have reduced the customers' ability to concentrate on the dialogue, resulting in lowered evaluations.

\section{CONCLUSION}
We developed a hybrid system of human-like and machine-like AR for DRC 2022 that has behaviors/utterances that are as smooth and natural as possible while focusing on asymmetric communication.
Our AR started with a machine-like character and transformed its voice tone and sentence expression in order to tailored to each customer.
This resulted in the customer's reaction being one of surprise or laughter.
Some tricky phrases for our experimental purpose, such as questions related to customer privacy and forceful requests for agreement, had a certain effect on customer's feeling of pleasure, reducing psychological distance and stress.
Although we still need further analyses, from an overall subjective evaluation after a 5-minute dialogue, the good scores for ``satisfaction with dialogue,'' ``trustworthiness,'' and ``robot recommendation effect'' suggest that our asymmetric communication system would be acceptable for humans nowadays.


\begin{thebibliography}{99}

\bibitem{Nishio 2007} 
S. Nishio, H. Ishiguro, N. Hagita, ``Geminoid: Teleoperated Android of an Existing Person,'' Humanoid Robots: New Developments, pp. 343--352, 2007.

\bibitem{Glas 2016} 
D. F. Glas, T. Minato, C. T. Ishi, T. Kawahara, H. Ishiguro, ``ERICA: The ERATO Intelligent Conversational Android,'' In Proc. of 25th IEEE International Symposium on Robot and Human Interactive Communication (RO-MAN), pp. 22--29, 2016.

\bibitem{Minato 2022} T. Minato, R. Higashinaka, H. Nishizaki, T. Nagai, K. Sakai, T. Funayama, ``Overview of Dialogue Robot Competition 2022,'' In Proc. of the Dialogue Robot Competition 2022, 2022.

\bibitem{Higashinaka 2022} 
R. Higashinaka, T. Minato, K. Sakai, T. Funayama, H. Nishizaki, T. Nagai, ``Dialogue Robot Competition for the Development of an Android Robot with Hospitality,'' In Proc. of IEEE 11th Global Conference on Consumer Electronics (GCCE2022), 2022.

\bibitem{Kawamoto 2022} 
M. Kawamoto, D. Kawakubo, H. Sugiyama, M. Shuzo, E. Maeda, ``Multi-modal Dialogue Strategy for Android Robots in Symbiotic Society,'' In Proc. of The 36th Annual Conference of the Japanese Society for Artificial Intelligence (JSAI 2022), 2N5-OS-7a-02, 2022. (in Japanese)

\bibitem{Bono 2015} M. Bono, ``Can a Robot Join an Idobata Kaigi?: A Fieldwork on Theatrical Creation of Daily Conversation,'' In Proc. of the 32nd Annual Meeting of the Japanese Cognitive Science Society, Vol. 22, No. 1, pp. 9--22, 2015. (in Japanese)

\bibitem{Kodama 2020} 
R. Tanaka, S. Kurohashi, ``Modeling and Utilizing User's Internal State in Movie Recommendation Dialogue,'' arXiv preprint arXiv:2012.03118, 2020.

\bibitem{Sugiyama 2021} 
H. Sugiyama, M. Mizukami, T. Arimoto, H. Narimatsu, Y. Chiba, H. Nakajima, T. Meguro, ``Empirical Analysis of Training Strategies of Transformer-Based Japanese Chit-Chat Systems,'' arXiv preprint arXiv:2109.05217, 2021.

\end{thebibliography}
\end{document}